\def\year{2022}\relax
\documentclass[letterpaper]{article} 
\usepackage{aaai22}  
\usepackage{times}  
\usepackage{helvet}  
\usepackage{courier}  
\usepackage[hyphens]{url}  
\usepackage{graphicx} 
\usepackage{natbib}  
\usepackage{caption} 
\DeclareCaptionStyle{ruled}{labelfont=normalfont,labelsep=colon,strut=off} 
\usepackage[ruled,linesnumbered]{algorithm2e}

\usepackage[table,dvipsnames]{xcolor}
\usepackage{subfigure}
\usepackage{color}
\usepackage{booktabs} 
\usepackage{amsmath}
\usepackage{bbm}
\usepackage{dsfont}
\usepackage{multirow, makecell}

\usepackage[normalem]{ulem}
\usepackage{amssymb}
\usepackage{multicol}

\usepackage{xspace}
\usepackage{paralist}
\usepackage{mdwlist}

\usepackage[switch]{lineno}

\newcommand{\method}{\textsc{EduCat}\xspace}
\newcommand{\timetravel}{\textsc{TimeTravel}\xspace}

\title{Unsupervised Editing for Counterfactual Stories}

\author {
    Jiangjie Chen\textsuperscript{\rm $\spadesuit\clubsuit$}\thanks{Work is done during internship at ByteDance AI Lab.},
    Chun Gan\textsuperscript{\rm $\diamondsuit$}\footnotemark[1], 
    Sijie Cheng\textsuperscript{\rm $\spadesuit$},
    Hao Zhou\textsuperscript{\rm $\clubsuit$}\thanks{Corresponding authors.},
    Yanghua Xiao\textsuperscript{\rm $\spadesuit\mathsection$}\footnotemark[2],
    Lei Li\textsuperscript{\rm $\heartsuit$}\thanks{Work is done while at ByteDance AI Lab.}
}
\affiliations {
    \textsuperscript{\rm $\spadesuit$}Shanghai Key Laboratory of Data Science, School of Computer Science, Fudan University\\
    \textsuperscript{\rm $\clubsuit$}ByteDance AI Lab
    \textsuperscript{\rm $\diamondsuit$}JD.com
    \textsuperscript{\rm $\heartsuit$}University of California, Santa Barbara\\
    \textsuperscript{\rm $\mathsection$}Fudan-Aishu Cognitive Intelligence Joint Research Center\\
    \texttt{\{jjchen19, sjcheng20, shawyh\}@fudan.edu.cn}, \\
    \texttt{cgan5@wisc.edu}, \ \ \texttt{zhouhao.nlp@bytedance.com}, \ \ 
    \texttt{lilei@cs.ucsb.edu}
}

\begin{document}

\maketitle

\begin{abstract}
Creating \textit{what-if} stories requires reasoning about prior statements and possible outcomes of the changed conditions.
One can easily generate coherent endings under new conditions, but it would be challenging for current systems to do it with minimal changes to the original story.
Therefore, one major challenge is the trade-off between generating a logical story and rewriting with minimal-edits.
In this paper, we propose \method, an editing-based unsupervised approach for counterfactual story rewriting.
\method includes a target position detection strategy based on estimating causal effects of the \textit{what-if} conditions, which keeps the causal invariant parts of the story.
\method then generates the stories under fluency, coherence and minimal-edits constraints.
We also propose a new metric to alleviate the shortcomings of current automatic metrics and better evaluate the trade-off.
We evaluate \method on a public counterfactual story rewriting benchmark.
Experiments show that \method achieves the best trade-off over unsupervised SOTA methods according to both automatic and human evaluation. 
The resources of \method are available at: \url{https://github.com/jiangjiechen/EDUCAT}.

\end{abstract}

\section{Introduction}
\label{sec:intro}

\emph{Counterfactual reasoning} is a hypothetical thinking process to assess possible outcomes by modifying certain prior conditions. 
It is commonly known as ``what-if'' analysis --- ``what will happen if \dots''.
It is a big challenge to build an intelligent system with counterfactual reasoning capabilities~\citep{pearl2009causality, pearl2018book}.
Counterfactual reasoning relies on the ability to find the \emph{causal invariance} in data, i.e. the factors held constant with the change of conditions in a series of events~\citep{sloman2004causal}.

\begin{figure}[ht]
    \centering
	\includegraphics[width=\linewidth]{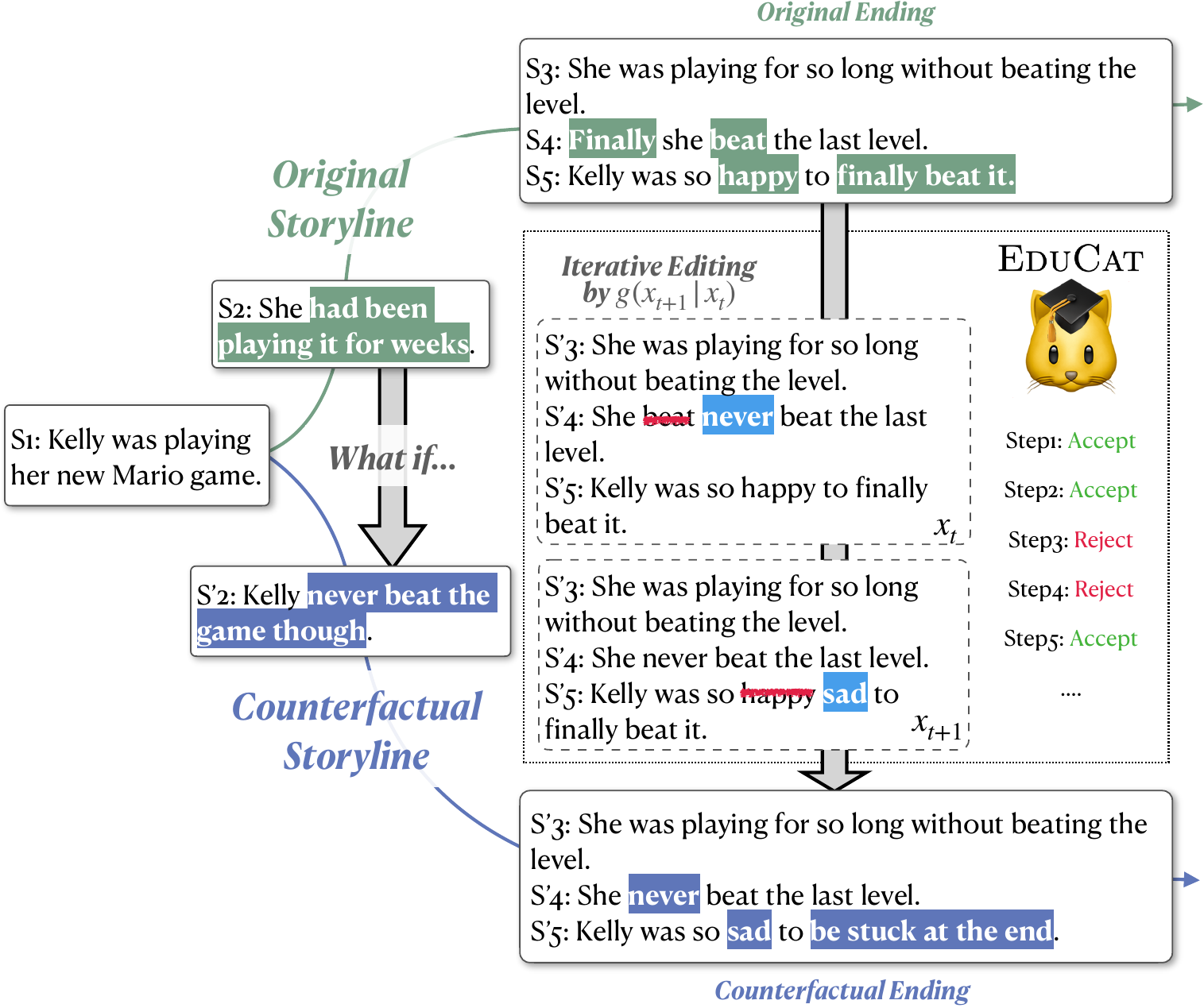}
    \caption{Counterfactual story rewriting example from the \timetravel \cite{qin-etal-2019-counterfactual} dataset. Our proposed \method iteratively edits the original ending to obtain new endings. }
    \label{fig:front}
\end{figure}

In this paper, we study \emph{unsupervised} counterfactual story rewriting, a concrete instance of counterfactual reasoning.
We focus on \emph{unsupervised} methods for this task, since humans do not need supervised learning to imagine alternative futures.
The task is to create plausible alternative endings given small modifications to the story context.

In this task, the major challenge is the trade-off between generating \emph{natural} stories and modifying the original text with \emph{minimal-edits}.
This requires finding the causal invariance in a story, i.e., invariant future events under the change of conditions.
Indeed, with a pre-trained language model (LM), it is relatively easy to generate fluent endings under new conditions with \emph{massive edits}.
However, difficulties arise when one has to perform accurate reasoning during modifying the ending \emph{minimally} while keeping it natural.

For example, in Figure \ref{fig:front}, what if Kelly played with the Mario game but \textit{never beat the game} (alter $s_2$ to $s_2'$)?
From human commonsense, one can easily create a plausible alternative story ending by making small edits that Kelly \textit{never} beat the last level rather than \textit{finally} beat it, and hence Kelly would be \textit{sad} instead of \textit{happy}.
In this case, the \textit{invariant} event is that Kelly still plays all levels until the last, but the variant event would be the consequence of the counterfactual intervention.
By identifying and keeping the invariant event, an ideal system can generate a plausible ending with few edits to the variant events.

Most of the existing methods \cite{li-etal-2018-generating,xu2018skeleton,guan2019story,guan2020knowledge} focus on the story generation in an auto-regressive manner.
These approaches keep the story logical mainly by exploiting the language modeling ability of LMs such as the GPTs \cite{radford2018improving,radford2019language,NEURIPS2020_1457c0d6}.
Few of them \cite{qin-etal-2019-counterfactual,qin2020backpropagation} deal with the reasoning ability in counterfactual text generation, which requires balancing between coherence and minimal-edits.
For example, \citet{qin2020backpropagation} propose to keep the balance by constraining the decoding on new endings with a sentence-level similarity scorer with the original ones.
However, LMs are known to be hard to control, often leading to over-editing.

In this paper, we propose \method, an \textbf{ED}iting-based \textbf{U}nsupervised \textbf{C}ounterfactual gener\textbf{AT}ion method for counterfactual story rewriting.
Given the original story and a modified condition statement, the challenge is to locate which part to retain (i.e. causal invariance) and which to modify (i.e. causal variance) while maintaining coherence to the context after editing.
Inspired by causal analysis research~\citep{hernan2004definition}, we quantify the potential outcome after intervention using the ratio between consistencies with the counterfactual and initial conditions, which can be computed by an off-the-shelf model.
\method employs a Markov chain Monte Carlo sampling framework \cite{metropolis1953equation} for unsupervised generation by iteratively generating token modifications \cite{miao2019cgmh}.
With desired properties and guidance from the estimated potential outcome, \method generates fluent and coherent alternative story endings with minimal edits.

The contributions of this work are as follows: 
\begin{itemize*}
    \item We first solve the counterfactual story rewriting task using unsupervised discrete editing method based on MCMC sampling.
    \item We draw inspiration from causal analysis and propose two counterfactual reasoning components that quantify the outcomes of context changes.
    \item We conduct experiments to verify that \method achieves the best trade-off between coherence and minimal-edits for unsupervised methods.
\end{itemize*}

\section{Task Formulation with Causal Model}
\label{sec:background}

In counterfactual story rewriting task, given a story consisting of a premise $z$, a story context $x$ and an ending $y$, we intervene by altering $x$ into a counterfactual context $x'$ and hope to predict new ending $y'$.

\begin{figure}[tp]
    \centering
	\includegraphics[width=\linewidth]{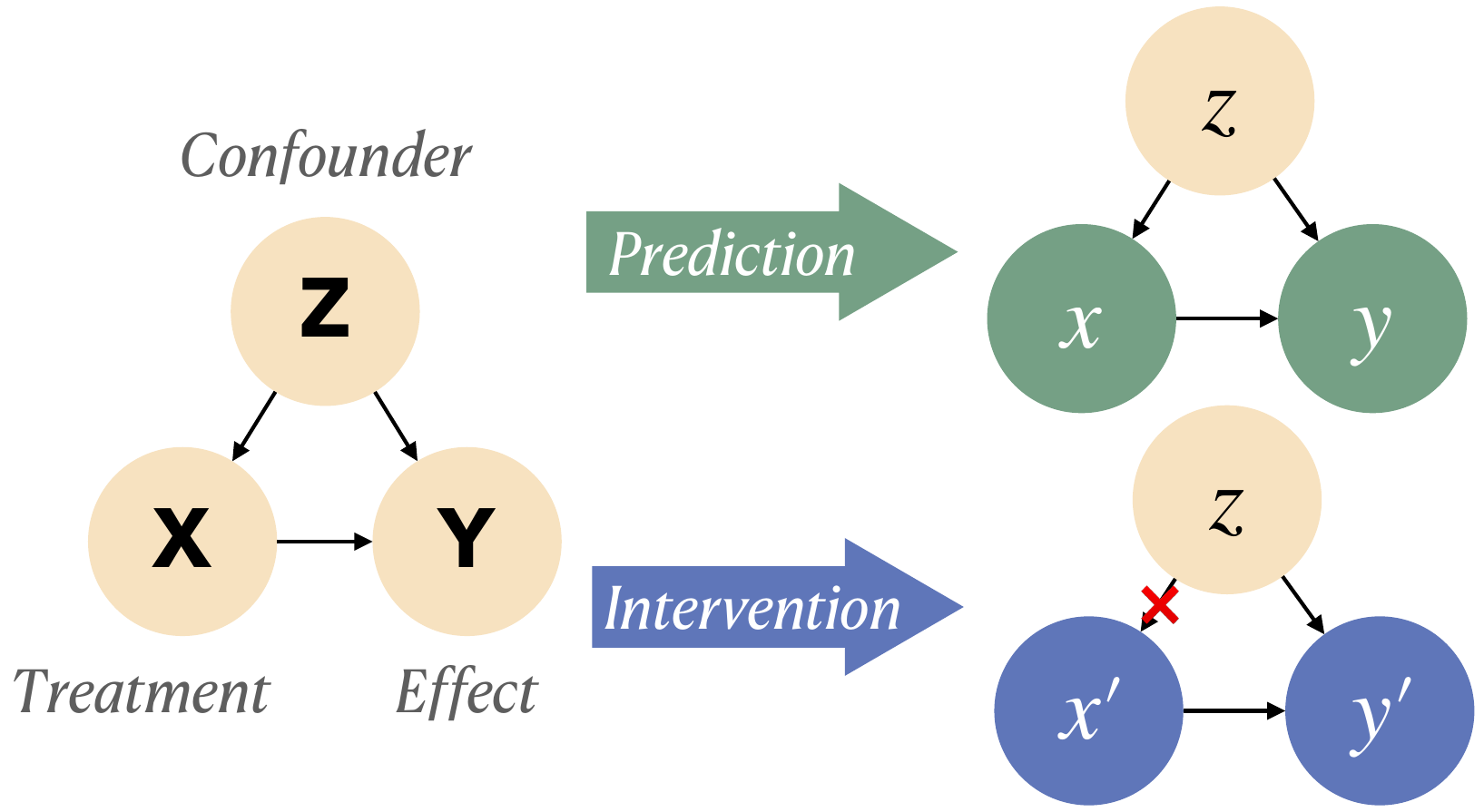}
    \caption{Formulating counterfactual story rewriting with intervention on causal model, where $z$ is the common premise of the story, $x, y$ denote the original story, and $x', y'$ are the counterfactual story.}
    \label{fig:init_count}
\end{figure}

This problem naturally fits to be formulated with a \textit{Causal Model}, a directed acyclic graph used to encode assumptions on the data generating process. 
As presented in the Figure \ref{fig:init_count}, the left part shows a simple example of a causal model with \textit{treatment} ($X$), \textit{effect} ($Y$) and \textit{confounder} ($Z$), respectively. 
In causal inference, a confounder is a random variable that influences both the treatment and effect variables, causing a spurious correlation \cite{pearl2009causality}.
Note that in this problem, $z$ consists of both observed confounder $s_1$ and unobserved commonsense knowledge, where the latter is very difficult to explicitly model.

The counterfactual inference can be formulated with a \textit{do}-operator.
As shown in Figure \ref{fig:init_count}, we can intervene on the $X$ variable by applying $\text{do}(X) = x'$ to set its value to the counterfactual without changing the rest. 
The arrow pointing from $Z$ to $X$ in the causal model is deleted since $X$ no longer depends on $Z$ after the intervention, resulting in a new graphical model. 
Consequently, the problem of counterfactual story generation can be formally restated as a counterfactual inference problem as follows: given $(z, x, y)$, what would the potential outcome of $y$ be if one changes the story context from $x$ to $x'$?

\section{Proposed Approach: \method}
\label{sec:method}

In this section, we present an overview and details of \method.
In general, the rewriting process works as follows:
starting with an original full story, \method performs the following procedures \textit{iteratively}:
\begin{enumerate}
    \item \emph{Conflict Detection}, it finds possible chunks in current story endings contradictory to counterfactual conditions;
    \item \emph{Edits Proposal}, it proposes an edited ending and decides its acceptance based on fluency and coherence scores.
\end{enumerate}
The above steps repeat multiple rounds.
Each proposal is either accepted or rejected based on desired properties $\pi(\mathrm{y})$, which is defined as the score product of each property score:
\begin{equation}
    \pi(\mathrm{y}) \propto \overbrace{\mathcal{X}_c^0(\mathrm{y}) \cdots \mathcal{X}_c^n(\mathrm{y})}^\text{Desired Properties}
\end{equation}
Finally, we pick the best one according to a ranking function as the output.
An illustrative example is given in Figure \ref{fig:front}.

However, the challenge remains for the quantification of these desired properties for ideal story rewriting.
Inspired by causal analysis research, we can quantitatively calculate the difference of story endings' quality given different conditions with the Causal Risk Ratio (CRR) \cite{hernan2004definition, hernan2020causal}. 
CRR is defined as follows:
\begin{equation}\label{CRR}
    \mathrm{CRR}  = \frac{\mathrm{P} (Y=y |\, \mathrm{do}(X = x'), Z=z)}{\mathrm{P}(Y=y |\, \mathrm{do}(X = x), Z=z)}
\end{equation}
The value goes up when the new ending is more consistent with the counterfactual condition. 
However, it is difficult to explicitly calculate both observed and unobserved confounders ($z^\star$) in $\mathrm{P}(Y=y|\, \mathrm{do}(X=x))$ as follows:
\begin{equation}
     \overbrace{\mathop{\sum}_{z^\star} \mathrm{P}(Y=y|X=x, Z=z^\star)\mathrm{P}(Z=z^\star)}^{\mathrm{P}(Y=y|\, \mathrm{do}(X=x))}
\end{equation}
We make a causal sufficiency assumption that only observed confounder ($z$) is considered:
\begin{equation}
    \mathrm{P}(Y=y|\, \mathrm{do}(X=x)) =\mathrm{P}(Y=y|X=x, Z=z)
\end{equation}
So CRR can be calculated by
\begin{equation}
    \mathrm{CRR}  = \frac{\mathrm{P} (Y=y |\, X = x', Z=z)}{\mathrm{P}(Y=y |\, X = x, Z=z)}
\end{equation}
In this way, we can roughly estimate the influence on possible endings brought by a changed condition.
Next, we will elaborate on the details of \method.

\subsection{Constrained Generation via MCMC}
\label{mcmc}

In \method, we direct the Markov chain Monte Carlo (MCMC) sampling process with counterfactual reasoning ability brought by conflict token detection and desired properties as sampling constraints.

\method directly samples from the sentence space with three local operations: token \textit{replacement}, \textit{deletion} and \textit{insertion}.
During sampling, after an edit position is found, the operation is randomly chosen with equal probability.
Finally, the proposed new sentence will either be accepted or rejected according to the \textit{acceptance rate} computed by desired properties $\pi(\mathrm{y})$.
The above process is repeated till convergence.

Specifically, Metropolis-Hasting sampling (MH) algorithm moves the current sentence $\mathrm{y}_{t}$ to the next sentence $\mathrm{y}_{t+1}$ by generating from the proposal distribution $g(\mathrm{y}_{t+1}|\mathrm{y}_{t})$ and accepting it based on an acceptance rate.
The sample distribution in MCMC will converge to the stationary distribution $\pi(\mathrm{y})$ in the Markov chain under mild conditions.
The acceptance rate $\alpha$ at the $t$-th iteration is defined as follows,
\begin{equation}
    \alpha(\mathrm{y}_{t+1}|\mathrm{y}_{t}) = \min\left\{1, \frac{\pi(\mathrm{y}_{t+1})^{1/T} g(\mathrm{y}_{t}|\mathrm{y}_{t+1})}{\pi(\mathrm{y}_{t})^{1/T} g(\mathrm{y}_{t+1}| \mathrm{y}_{t})}\right\}
\end{equation}
$T$ is a temperature controlled by a cooling schedule \cite{andrieu2003introduction} ($T = 0.95^{\lfloor \frac{t}{5} \rfloor}$ in our implementation.) 

Next, we will describe in detail the design of stationary distribution $\pi(\mathrm{y})$ ($\mathsection$\ref{sec:constraints}) and transition proposal distribution $g(\mathrm{y}_{t+1}|\mathrm{y}_{t})$ ($\mathsection$\ref{sec:proposal}).

\subsection{Desired Properties for Story Rewriting}
\label{sec:constraints}
Aside from the basic fluency property, the original CGMH framework is designed with properties such as similarity and keywords constraints. 
These simple properties cannot direct the sampling with counterfactual reasoning ability.
Instead, we want the generated new endings to be not only \textit{fluent} in terms of storytelling, but also logically \textit{coherent} with $X'$ instead of $X$.
In \method, we define two score functions in story rewriting, namely, a fluency score function $\mathcal{X}_{\mathrm{LM}}$ and a coherence score function $\mathcal{X}_{\mathrm{Coh}}$.
Thus, the stationary distribution $\pi(\mathrm{y})$ is defined as the product of fluency score and the coherence score as follows:
\begin{equation}
    \pi(\mathrm{y}) \propto \mathcal{X}_\mathrm{LM}(\mathrm{y})\cdots \mathcal{X}_\mathrm{Coh}(\mathrm{y})
\end{equation}

\subsubsection{Fluency Score}
We compute the probability of the generated ending based on a pre-trained language model, e.g. GPT-2 \cite{radford2019language}.
This is important and in line with previous work to guarantee the fluency and readability of the generated sentence.
The likelihood is computed autoregressively as:
\begin{equation}
    \mathcal{X}_\mathrm{LM}(y^*) = \prod_{i=1}^N P_\mathrm{LM}(y^*_i | z, x', y^*_{<i}).
\end{equation}
We denote $y^*$ as the proposed ending at the current stage, and $y_i^*$ as the $i$-th token in the ending.

\subsubsection{Coherence Score}

Intuitively, we want to punish proposed endings contradictory to the counterfactual conditions but consistent with the initial ones.
Therefore, the purpose of coherence score function $\mathcal{X}_\mathrm{Coh}$ is to encourage the model to rewrite the original endings.
The value of $\mathcal{X}_\mathrm{Coh}$ should be larger than $1$ if the generated ending is more causally related to counterfactual context than the initial one.
Inspired by the definition of Causal Risk Ratio, the coherence score function $\mathcal{X}_\mathrm{Coh}$ is defined as follows:
\begin{equation}
\mathcal{X}_\mathrm{Coh}(y^*) = \frac{P_\mathrm{Coh}(Y = y^* |\,z, x')}{P_\mathrm{Coh}(Y = y^* |\,z, x)}
\end{equation}
where the formulation for $P_\mathrm{Coh}$ is fit for any model for quantification that measures the coherence between an ending and a story context.
In our implementation, we employ conditional sentence probability calculated by a pre-trained language model (e.g., a GPT-2) to measure the coherence within a story in an unsupervised way.
Note that we hope to solve this task in an unsupervised way.
But $P_\mathrm{Coh}$ is fully extendable for better story coherence checking models.

\subsection{Editing Proposal Design}
\label{sec:proposal}

Regularized by the desired properties, we can make editing proposals by solving two questions: 1) \textit{Where to edit?} and 2) \textit{Edit with what?}

\subsubsection{Where to Edit: Conflict Detection}
\label{posfinding}
It is critical to know where to edit the original stories to write natural counterfactual stories with only minimal edits.
Namely, we need to identify tokens that contradict with the counterfactual context \cite{Hao_Pang_Lan_Wang_Guo_Cheng_2021}.
Meanwhile, causal invariant information is kept in the unchanged tokens.

Also inspired by the calculation of Causal Risk Ratio, we estimate the potential outcome of changing the contexts to find the most likely contradictory tokens.
Let $y^*$ be the current ending to edit (initialized with $y$) and $y^*_i$ be the tokens, we define the conflicting probability $P_\mathrm{cf}(y^*_i)$ on the $i$-th token in $y^*$ as follows,
\begin{equation}
\label{ttf}
P_\mathrm{cf}(y^*_i) = \mathrm{softmax}( \frac{P_\mathrm{LM}(y^*_i | z, x, \, y^*_{<i})}{P_\mathrm{LM}(y^*_i |z, x', \, y^*_{<i})})
\end{equation}
The token-level likelihood is computed via a language model.
According to the definition, $P_\mathrm{cf}(y^*_i)$ is larger if $y^*_i$ is more causally related to the initial context than the counterfactual one. 
Those tokens are more likely to contradict with counterfactual conditions at each iteration.
They should have a higher priority to be edited.

\subsubsection{Edit with What: Modification Action}
We randomly sample from three token-level modification actions (replacement, deletion, and insertion) with equal probability to find what to use to edit the endings given editing positions.

Let $\mathrm{y}_{t}$ be the current sentence, the proposal distribution is defined as $g(\mathrm{y}_{t+1} | \mathrm{y}_{t})$.
The expectation of transition proposal from $\mathrm{y}_{t}$ to $\mathrm{y}_{t+1}$ is given by 
\begin{equation}
    g(\mathrm{y}_{t+1} | \mathrm{y}_{t}) = \frac{1}{3}\sum_{\mathrm{op} \in \{\mathrm{r,d,i}\}}g_\mathrm{op}(\mathrm{y}_{t+1}|\mathrm{y}_{t})
\end{equation}
where $g_\mathrm{r}$, $g_\mathrm{d}$, $g_\mathrm{i}$ correspond to the replacement, deletion and insertion proposals, respectively.
For \textit{replacement}, let $\mathrm{y}_{t} = [w_1, \dots, w_m, \dots, w_n]$, the replacement action replaces the token $w_m$ with $w^c$, where $w^c$ is sampled from a pre-selected candidate set $\mathcal{Q}$.
Let $\mathrm{y}_{t+1}=[w_1, \dots, w^c, \dots, w_n]$, then the proposal for replacement is
\begin{equation}
    \begin{aligned}
    \label{replace}
    &g_{r}(\mathrm{y}_{t+1} | \mathrm{y}_{t})=\mathds{1}(w^c \in \mathcal{Q}) \cdot P_{\text{MLM}}(w_m^*=w^c|\mathrm x_{-m})
    \end{aligned}
\end{equation}
Here $\mathds{1}(w^c \in \mathcal{Q})$ is the indicator function which equals $1$ if $w^c \in \mathcal{Q}$ and $0$ otherwise. $P_{\text{MLM}}(w_m^*=w^c|\mathrm x_{-m})$ is the probability of the selected token given the rest of the sentence $x_{-m}$. 
It is computed using a masked language model (MLM), e.g. BERT \cite{devlin2019bertpo} or RoBERTa \cite{liu2019roberta}. 

The transition function for \textit{deletion} is rather simple: $g_d(\mathrm{y}_{t+1} | \mathrm{y}_{t}) = 1$ if and only if $\mathrm{y}_{t+1} = [w_1,\dots,w_{m-1},w_{m+1}, \dots, w_n]$, and 0 for others.
The \textit{insertion} operation consists of two steps. 
First, a mask token is inserted into the position and then a replacement operation is performed on the inserted token.

\section{Experiments}
\label{sec:experiment}
\subsection{Experimental Setup}

\paragraph{Dataset}
We experiment \method on \timetravel \cite{qin-etal-2019-counterfactual}, a standard counterfactual story rewriting dataset.
\timetravel is built on ROCStories \cite{DBLP:journals/corr/MostafazadehCHP16}, which consists of a large set of five-sentence stories $S=s_{1:5}$.
The first sentence $s_1$ denotes the premise of a story, $s_2$ sets up the initial context, and the last three sentences $s_{3:5}$ are the story endings.
Using causal language we described above, $s_1$, $s_2$, $s_{3:5}$ correspond to $Z=z$, $X=x$, $Y=y$, respectively.
In \timetravel, the initial context was rewritten by humans into a counterfactual context $s_2'$, followed with edited endings $s_{3:5}'$.
They correspond to $X=x'$ and $Y=y'$ in the causal graphical model.
As \method is unsupervised and thus does not need training, we run \method directly on the test set.

The statistics of \timetravel are reported in Table \ref{tab:dataset}.
Only part of the training set is annotated with the edited endings.
Each sample in the development and test set is annotated with 3 and 4 rewritten endings respectively, which explains the difference between \# of $x'$ and \# of $y'$ in the development and test set in Table \ref{tab:dataset}.
Note that the \emph{fourth edited ending} in test set is not included in evaluation as ground truth ending, but \emph{only} serves as human baseline.

\begin{table}[tp]
    \centering
    \begin{tabular}{cccc}
        \toprule
         & \textbf{Train} & \textbf{Dev} & \textbf{Test} \\
        \midrule
        \# counterfactual context ($x'$) & 96,867 & 1,871 & 1,871 \\
        \# edited endings ($y'$) & 16,752 & 5,613 & 7,484 \\
        \bottomrule
    \end{tabular}
    \caption{Statistics of \timetravel dataset.}
    \label{tab:dataset}
\end{table}

\paragraph{Baselines}
Following previous work, we categorize the baselines into three classes: 
\begin{inparaenum}[\it 1)]
    \item \emph{Unsupervised zero-shot baselines}, with only off-the-shelf pre-trained models for generation, including pre-trained GPT-2 (generating with $s_1, s_2'$) and \textsc{Delorean} \cite{qin2020backpropagation}. 
    Moreover, in comparisons with unsupervised editing-based methods, we add CGMH \cite{miao2019cgmh}, which is \method without conflict detection and coherence score;
    \item \emph{Unsupervised training baselines}, GPT-2 + \texttt{Recon+CF} \cite{qin-etal-2019-counterfactual}, which is \emph{trained} with domain data $S$ and $<s_1, s_{2}'>$ (i.e. without $s_{3:5}'$);
    \item \emph{Supervised training baselines}, with a GPT-2 + \texttt{SUP} \cite{qin-etal-2019-counterfactual} \emph{trained} for predicting $s'_{3:5}$ from $S$ and $s_2'$ in the form of $<S, \text{[SEP]}, s_1, s_2'>$.
\end{inparaenum}

Note that in our paper, we aim at using only off-the-shelf pre-trained models for story rewriting, which makes the previous SOTA method \textsc{Delorean} our major baseline.
\textsc{Delorean} iteratively revises the generated tokens by updating their hidden representations during decoding.
The update is constrained by minimizing the sentence-level KL divergence between the generated and original endings, followed by a BERT to re-rank the generated candidates with the next sentence prediction task.

\paragraph{Implementation Details}
All of the pre-trained checkpoints are inherited from the implementations of Huggingface \cite{wolf-etal-2020-transformers}.
Consistent with previous work, we adopt GPT-2, Medium (24 layers) or Small (12 layers), for causal language modeling.
We use pre-trained RoBERTa-base as the unsupervised masked language model for token proposal.
We keep the first 100 tokens MLM predicts as candidates. We randomly sample one token as the proposed token based on normalized probabilities.
In the experiments, we run \method and its variants for 100 steps.

\subsection{Evaluation Metrics}

\subsubsection{Automatic Evaluation Metrics}

Following previous work, we adopt BLEU-4 \cite{papineni2002bleu} and \textsc{BERTScore} \cite{DBLP:conf/iclr/ZhangKWWA20} as automatic metrics, which are referenced metrics.
Given ground-truth endings and the generated endings, BLEU computes the number of overlapping n-grams, and \textsc{BERTScore} computes their semantic similarity using BERT.
As reported in \citet{qin-etal-2019-counterfactual}, BLEU measures the \textit{minimal-edits} property well, but correlates poorly with human judgements w.r.t. \textit{coherence}.

For assessing the \textit{coherence} with the counterfactual conditions, we propose a simple, unreferenced, and model-based metric \textsc{EntScore} (\textsc{EntS}).
Inspired by researches on natural language inference \cite{kang2018adventureat,dziri-etal-2019-evaluating},
we fine-tune a RoBERTa (base or large) with \textit{binary} classification objective to check whether a story context entails a story ending.
We use 28,363 stories with annotated edited endings in \timetravel to train the metric, leading to 113,452 training samples, i.e., $x'$ contradicts with $y$ but entails by $y'$ and $x$ contradicts with $y'$ but entails $y$.
The best metrics achieve the F1 scores of 73.07 (base) and 81.64 (large) in the test set.
We take the predicted probability of whether an ending is entailed by the counterfactual context as the output of \textsc{EntScore}.

To better evaluate the subtle trade-off in this task, we calculate a \emph{harmonic mean} of \textsc{EntScore} and BLEU to represent the trade-off between coherence and minimal-edits, defined as $\textsc{HMean} = \frac{2\cdot \text{BLEU}\cdot \textsc{EntS}}{\text{BLEU} + \textsc{EntS}}$.

\begin{table}[tp]
    \centering 
    \small 
    \begin{tabular}{cccc}
        \toprule
        \textbf{Metrics} & \textbf{Pearson's $r$} & \textbf{Spearman's $\rho$} & \textbf{Kendall's $\tau$}  \\
        \midrule
        BLEU & 0.2619 & 0.2454 & 0.1758 \\
        \textsc{BERTScore} & 0.3252 & 0.3332 & 0.2385 \\
        \textsc{EntS} (base) & 0.3937 & 0.3973 & 0.2865 \\
        \textsc{EntS} (large) & 0.4685 & 0.4732 & 0.3389 \\
        \textsc{HMean} (large) & \textbf{0.4995} & \textbf{0.4996} & \textbf{0.3662} \\
        \bottomrule
    \end{tabular}
    \caption{The correlation between automatic metrics and human judgements in coherence. \textsc{HMean} is the harmonic mean between \textsc{EntS} (large) and BLEU. All of these numbers are statistically significant at $p < 0.01$. }
    \label{tab:corr}
\end{table}

\subsubsection{Human Evaluation Metrics}
 
We also conduct human evaluation to compensate for these automatic metrics and assess their ability for this task.
Following \citet{qin2020backpropagation}, our human evaluation mainly focuses on two primary criteria: i) \textit{coherence}, the logical consistency between the counterfactual context ($s_1, s_2'$) and generated endings, and ii) \textit{minimal-edits}, the extent of minimal revision between two endings. 
We calculate the pairwise comparison as human metrics.
Annotators are asked to score from 0 to 3 and choose the better one or both between two generated outputs from \method and baselines without knowledge of their origins.
We arrange a training session before annotation session, where the annotators annotate some cases and resolve their disputes through discussion.
Then, we randomly select 100 samples from the test set. 
Each sample was rated by three graduate students, paid with local minimum wage.\footnote{They reach fair inter-rater agreement with Fleiss' $\kappa=0.345$ in annotation session.}
The final decision is made based on the majority vote.

\subsubsection{Human Correlation with Metrics}

Before automatic evaluation, we show the ability of these automatic metrics by performing correlation analysis using the scores produced by human annotators on the generated endings.
We calculate three coefficients, including Pearson's $r$, Spearman's $\rho$ and Kendall's $\tau$.
Pearson's $r$ measures linear correlation, and the latter two measure monotonic correlation, where Spearman's $\rho$ is more sensitive to abnormal values.
According to Table \ref{tab:corr}, \textsc{HMean} proves to be the best metric among them in terms of correlation with human judgements for this task, which is also our primary metric in the experiments.

\subsection{Results}

\begin{table}[tp]
    \centering
    \begin{tabular}{lcccc}
        \toprule
        \textbf{Method} & \textbf{BLEU} & \textbf{BERT} & \textbf{\textsc{EntS}$_l$} & \textbf{\textsc{HMean}} \\
        \midrule
        \rowcolor[gray]{0.95} \multicolumn{5}{c}{\textit{Supervised Training}}\\
        GPT-2$_M$ + \texttt{SUP} & 76.35 & 81.72 & 35.06 & 48.05 \\
        \rowcolor[gray]{0.95} \multicolumn{5}{c}{\textit{Unsupervised Training}}\\
        GPT-2$_M$ + \texttt{FT} & 3.90 & 53.00 & 52.77 & 7.26 \\
        \texttt{Recon+CF} & 76.37 & 80.20 & 18.00 & 29.13 \\
        \rowcolor[gray]{0.95} \multicolumn{5}{c}{\textit{Off-the-shelf Pre-trained Models}} \\
        GPT-2$_{M}$ & 1.39 & 47.13 & \textbf{54.21} & 2.71 \\
        \textsc{Delorean} &  23.89 & 59.88 & 51.40  & 32.62 \\
        CGMH & 41.34 & 73.82 & 29.80 & 34.63 \\
        \method & \textbf{44.05} & \textbf{74.06} & 32.28 & \textbf{37.26} \\
        \midrule
        Human & 64.76 & 78.82 & 80.56 & 71.80 \\
        \bottomrule
    \end{tabular}
    \caption{Automatic evaluation results in the test set of \timetravel. These methods use GPT-2$_M$ by default. \textsc{EntS}$_l$ is short for \textsc{EntScore} (large).}
    \label{tab:auto}
\end{table}

\subsubsection{Automatic Evaluation}

Table \ref{tab:auto} shows our results w.r.t. automatic metrics.
In general, we observe that BLEU and \textsc{EntScore} indicate the trade-off between minimal edits and coherence in this task.
Models that generate coherent endings can also cause excessive edits.
Among them, \method achieves the best trade-off in terms of \textsc{HMean}, which is also the metric that has the best correlation with human judgements, as shown in Table \ref{tab:corr}.

For supervised and unsupervised training methods, we find \texttt{Recon+CF} scores high on BLEU and \textsc{BERTScore} but low on \textsc{EntScore}, suggesting that the endings it generates are not coherent with counterfactual contexts but paraphrased from original endings \cite{qin-etal-2019-counterfactual}.
Moreover, the gap remains between supervised methods and unsupervised ones.

Interestingly, zero-shot GPT-2$_M$ and \textsc{Delorean} perform very well in \textsc{EntScore} but poorly on BLEU and \textsc{BERTScore}.
\textsc{EntScore} draws the decision boundary based on the change of conditions ($s_2$, $s_2'$).
Therefore, as long as the ending follows the counterfactual condition, where large-scale language models such as GPT-2 excel, \textsc{EntScore} will produce a high score. 
Zero-shot GPT-2$_M$ does not constrain the generation on minimal-edits to the original endings and hallucinates from the original story during the generation. 
Hence, it generates fluent endings thanks to the language modeling ability of GPT-2 with \emph{over-editing}.
The same is true for \textsc{Delorean}, but it alleviates this problem by constraining on the KL-divergence with original endings.
Indeed, it is easy to generate coherent endings with \emph{massive edits}, as even a zero-shot GPT-2 can achieve a high score in coherence.
However, this task puts forward higher demands on the model's ability to do it under \emph{minimal edits} to find the causal invariance.

\subsubsection{Human Evaluation}

We first show manual evaluation results in Table \ref{tab:pairwise}.
In general, \method outperforms CGMH and \textsc{Delorean} w.r.t. \textit{coherence} and \textit{minimal-edits}.
\method achieves the similar results with CGMH on min-edits because they run for the same editing steps.

We observe in Table \ref{tab:pairwise} that \textsc{Delorean} is outperformed by \method in coherence. 
This seems contradictory with the automatic evaluation results reported before in terms of \textsc{EntScore}.
The possible reasons are two-fold.
First, \textsc{EntScore} is trained only with a simple discriminative classification objective, and is therefore sensitive to the change in the altered condition ($x \rightarrow x'$).
However, the coherence to the premise is also important to find causal invariance in counterfactual reasoning.
Not only do we focus on the coherence of the new story, we also highlight the minimal effort to make it happen.
And, \textsc{Delorean}, like GPT-2$_M$, is easy to hallucinate from the original story line.
Second, humans enjoy great ability in making up ``headcanons'' in their minds to connect two events, thus small but critical edits can still result in a logical ending to a human mind.

\begin{table}[t]
    \centering 
    \small
    \begin{tabular}{lccc}
    \toprule
    \multirow{2}{*}{\textbf{Methods}} & \multicolumn{3}{c}{\textbf{Coherence}}  \\ 
    \cmidrule{2-4}
     & \textbf{Win}  & \textbf{Tie} & \textbf{Lose}  \\ 
    \midrule
    \method vs. \textsc{Delorean} & 45\% & 32\%  & 23\%  \\
    \method vs. CGMH  &  32\% & 51\% & 17\% \\
    \method vs. Human   &  12\%  &  24\%  & 64\%  \\
    \midrule
    & \multicolumn{3}{c}{\textbf{Min-edits}}\\
    \cmidrule{2-4}
    \method vs. \textsc{Delorean}  & 64\% & 27\%  & 9\% \\ 
    \method vs. CGMH  & 26\% & 49\% & 25\% \\      
    \method vs. Human & 16\% & 40\% & 44\% \\
    \bottomrule
    \end{tabular}
    \caption{Manual evaluation results, with scores denoting the percentage of \textit{Win}, \textit{Lose} or \textit{Tie} when comparing \method with baselines.}
    \label{tab:pairwise}
\end{table}

\subsubsection{Ablation Study}

We perform an ablation study for the proposed modules.
We find both components are beneficial to this task according to Table \ref{tab:ablation} in all metrics.
Even with smaller GPT-2$_S$ as the backbone causal language model, \method still outperforms unsupervised baselines.

In particular, we find a considerable performance drop in BLEU and \textsc{EntScore} for \method without conflict detection module.
This result suggests that random edit token finding is inefficient to find the causal invariance.
So the method prefers the editing actions that generate fluent endings instead of ones that balance the trade-off well, which puts forth higher demands to the system.

We observe a mild performance boost in the trade-off (\textsc{HMean}) by introducing $\mathcal{X}_\mathrm{Coh}$ with unsupervised conditional sentence probability as the coherence function $P_\mathrm{Coh}$.
What if \method has more powerful coherence guidance from $\mathcal{X}_\mathrm{Coh}$? 
To test the limit of our method, we also upgrade $\mathcal{X}_\mathrm{Coh}$ by directly replacing the original $P_\mathrm{Coh}$ with \textsc{EntScore} (base), since the unsupervised sentence probability as the coherence measurement might be weak for the story domain.
Results indicate that using \textsc{EntScore} in $\mathcal{X}_\mathrm{Coh}$ leads to a clear boost in coherence (+30.20\% in \textsc{EntScore}) and the trade-off (+14.95\% in \textsc{HMean}).
This shows the potential of \method framework for this task given a robust discriminator, which is also similar to the benefits of a strong reward function in reinforcement learning.
Nevertheless, to keep this method solely unsupervised with only off-the-shelf models, we claim scores achieved by \method with the original $\mathcal{X}_\mathrm{Coh}$ as our major results, but with much room for improvement.

\begin{table}[tp]
    \centering
    \small
    \begin{tabular}{lcccc}
        \toprule
        \textbf{Ablation} & \textbf{BLEU} & \textbf{BERT} & \textbf{\textsc{EntS}$_l$} & \textbf{\textsc{HMean}}\\
        \midrule
        \method (GPT-2$_{S}$) & 39.82 & 72.35 & 31.72  & 35.31 \\
        \method (GPT-2$_M$) & 44.05 & 74.06 & 32.28 & 37.26 \\
        -- $\mathcal{X}_\mathrm{Coh}$ & \textbf{44.20}  & \textbf{74.27} & 31.44 & 36.74 \\
        -- \textit{conflict detection} & 40.96 & 73.61 & 30.79 & 35.16 \\
        -- \textit{both} & 41.34 & 73.82 & 29.80 & 34.63 \\
        + $\mathcal{X}_\mathrm{Coh}$ w/ \textsc{EntS}$_b$ & 43.65 & 74.09 & \textbf{42.03} & \textbf{42.83} \\
        \bottomrule
    \end{tabular}
    \caption{Ablation study of \method in terms of conflict detection module and coherence score $\mathcal{X}_\mathrm{Coh}$. We also change the $P_\mathrm{Coh}$ in $\mathcal{X}_\mathrm{Coh}$ to the trained discriminative metric \textsc{EntScore}.}
    \label{tab:ablation}
\end{table}

\subsection{Case Study}
\label{casestudy}

\begin{figure*}[t]
    \centering
	\includegraphics[width=\linewidth]{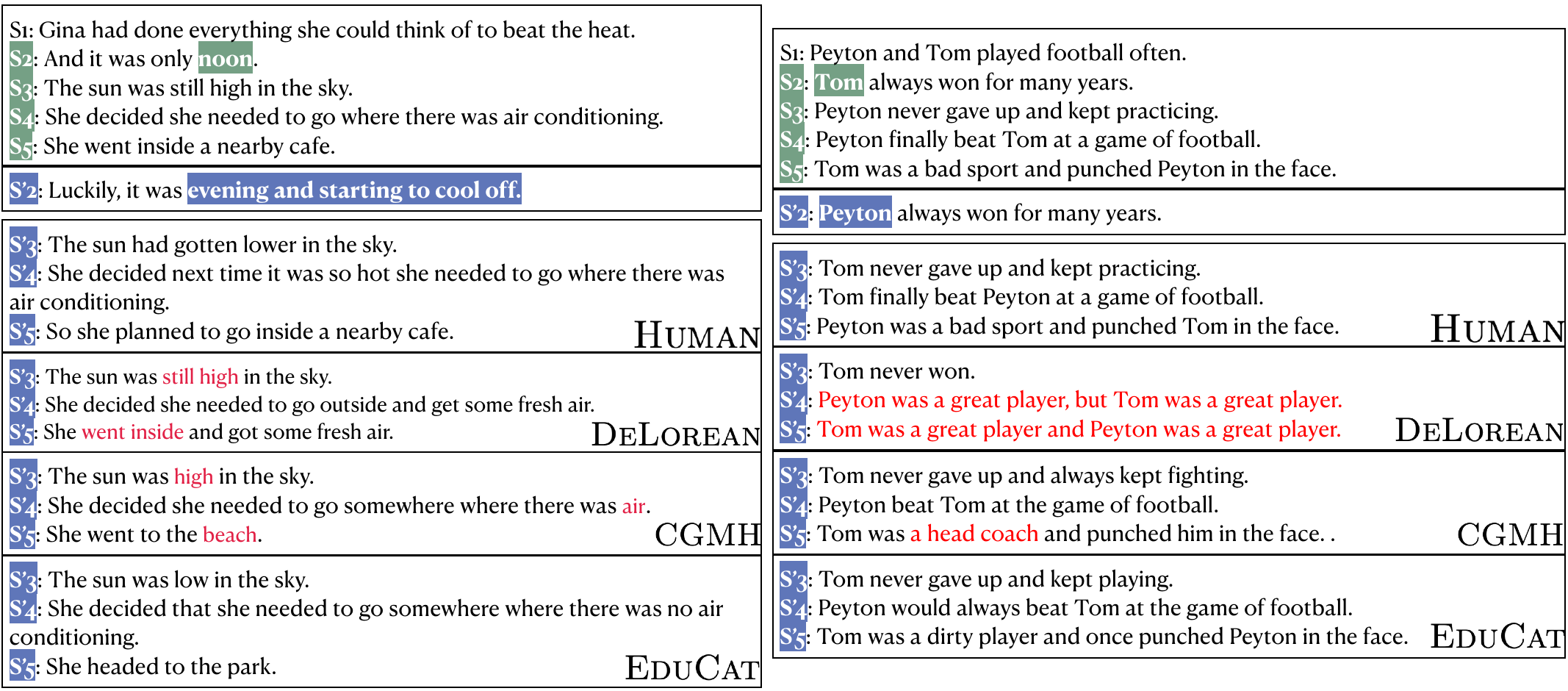}
    \caption{Two samples from the test set of \timetravel. We present the predictions of \method and baselines. Text in \textcolor{red}{red} denotes the \textcolor{red}{mistakes} these models make.}
    \label{fig:case}
\end{figure*}

Finally, we show some of the samples produced by \method against baselines in Figure \ref{fig:case} to make an intuitive comparison and explore our method's limitations.
Although \textsc{Delorean} also generates fluent counterfactual stories, it struggles at maintaining the balance between minimal-edits and logical consistency to the counterfactual context, and makes massive edits.
In contrast, the discrete editing strategy \method works far better than the gradient update-based method in \textsc{Delorean} in terms of minimal edits.

In both cases, \method and CGMH conduct a handful of edits to the original endings and yield fluent endings.
In the first one, \method makes crucial and logical lexical edits, e.g., the sun's position should be \textit{low} since it is evening in the altered condition $s_2'$, while CGMH and \textsc{Delorean} do not.
\method shows some commonsense knowledge, as one needs no air conditioning as the weather was starting to cool off, and \textit{park} is a good place to go in the evening (maybe for a walk).
In the second one, \textsc{Delorean} does not generate valid story endings.
CGMH makes mistakes by changing ``bad sport'' to ``head coach'', whereas \method paraphrases it to ``dirty player''.

\section{Related Work}
\label{sec:related}
\paragraph{Constrained Text Generation}

Many research efforts have been made to control the generation with various desired properties.
Most studies \cite{hu2018deep,tan-etal-2020-summarizing} train supervised models to inject constraints into generation.
In this work, we focus on unsupervised constrained generation, which is much more difficult.
Recent unsupervised generation relies heavily on pre-trained language models (PLMs) \cite{radford2019language,keskar2019ctrl}.
\citet{dathathri2019plug} control the generation using an external attribute model that affects token decoding through back-propagation.
\citet{qin2020backpropagation} adopt this idea and adjust for this task by optimizing the sentence generation as a whole through iterative forward and backward passes.

Another line of unsupervised constrained generation is search-based methods, including methods with constrained beam search \cite{hokamp-liu-2017-lexically,lu2020neurologic} and stochastic search.
The former line of work is restricted to lexical constraints, while the latter is more extendable.
\citet{miao2019cgmh} first introduce Metropolis-Hastings sampling into text generation and constrain the generation with stationary distributions.
\citet{zhang2020language} extend CGMH by designing combinatorial constraints.
\citet{liu-etal-2020-unsupervised} model the constraint generation as a discrete optimization problem, which is solved with simulated annealing.
To find edit positions, \citet{sha2020gradient} define differentiable score functions and use gradients to find edit positions and sample actions, while \citet{He_Li_2021} train a position finding classifier with XL-Net \cite{yang2019xlnet} for lexically constrained sentence generation.
In this paper, we mainly explore this line of work to non-monotonic reasoning and generation tasks with insights from causal analysis.

\paragraph{Causal Inference and NLP}
There is a recent surge of interest in how NLP methodology can evaluate and estimate causal effects and how causal inference can enhance current natural language understanding and generation.
Researchers have studied how text can be used as a mediator, confounder, treatment, or outcome \cite{grimmer2017estimating, wood2018challenges,wu2020biased,feder2021causal} to estimate causal effect under different contexts such as gender bias, etc. 
Another line of research attempts to equip the current text generation mechanism with counterfactual reasoning ability. 
For instance, \citet{kaushik2020learning,zeng2020counterfactual} augment existing datasets to include counterfactual samples and demonstrate better out of domain generalization ability on tasks as sentimental classification, NER, etc. 
In terms of work more related to ours \cite{zhu2020counterfactual,qin-etal-2019-counterfactual,qin2020backpropagation}, they explored the counterfactual text generation tasks such as counterfactual dialogue and story generation.
Our work adapts idea from both lines of researches.

\section{Conclusion and Future Work}
\label{sec:conclusion}
In this paper, we aim to balance the trade-off between logic and minimal-edits in order to detect causal invariance in the story rewriting task, which demands causal reasoning skills.
We propose \method, an editing-based unsupervised counterfactual story rewriter using MCMC sampling.
For detecting causal invariance, \method is equipped with the ability of conflict detection and scores for coherence to control the edit proposals based on causal risk ratio, a measure of causal effects.
Experiments on the \timetravel dataset show that \method substantially outperforms unsupervised SOTA methods in both automatic and human evaluation metrics, indicating the superiority of editing-based methods in this task.
Further ablation study stresses the importance of the proposed causal reasoning components.
Although this work makes an attempt on automatic evaluation of this task by proposing \textsc{EntScore}, we highlight that future research should prioritize on the automatic metrics for this task, especially for unreferenced metrics.

\section*{Acknowledgements}
We thank Changzhi Sun, Xinbo Zhang, Yuxuan Song, Chao Wang and the anonymous reviewers for the discussions for the manuscript.
We also thank Lianhui Qin for providing baseline results.
This work was supported by National Key Research and Development Project (No. 2020AAA0109302), Shanghai Science and Technology Innovation Action Plan (No.19511120400) and Shanghai Municipal Science and Technology Major Project (No.2021SHZDZX0103).

\bibliography{aaai22}

\end{document}